# A Novel Ensemble-Based Deep Learning Model with Explainable AI for Accurate Kidney Disease Diagnosis

Md. Arifuzzaman
Department of CSE
Leading University
Sylhet, Bangladesh
arif_cse@lus.ac.bd

Iftekhar Ahmed
Department of CSE
Leading University
Sylhet, Bangladesh
iftekharifat007@gmail.com

Md. Jalal Uddin Chowdhury
Department of CSE
Leading University
Sylhet, Bangladesh
jalal_cse@lus.ac.bd

Shadman Sakib
Department of Information Systems
University of Maryland
Baltimore, United States
sakibshadman15@gmail.com

Mohammad Shoaib Rahman
Department of CSE
Leading University
Sylhet, Bangladesh
shoaib_cse@lus.ac.bd

Md. Ebrahim Hossain
Department of CSE
Leading University
Sylhet, Bangladesh
ebrahim.cse@lus.ac.bd

Shakib Absar
Department of CSE
Leading University
Sylhet, Bangladesh
sabsar42@gmail.com

*Abstract*—Chronic Kidney Disease (CKD) represents a significant global health challenge, characterized by the progressive decline in renal function, leading to the accumulation of waste products and disruptions in fluid balance within the body. Given its pervasive impact on public health, there is a pressing need for effective diagnostic tools to enable timely intervention. Our study delves into the application of cutting-edge transfer learning models for the early detection of CKD. Leveraging a comprehensive and publicly available dataset, we meticulously evaluate the performance of several state-of-the-art models, including EfficientNetV2, InceptionNetV2, MobileNetV2, and the Vision Transformer (ViT) technique. Remarkably, our analysis demonstrates superior accuracy rates, surpassing the 90% threshold with MobileNetV2 and achieving 91.5% accuracy with ViT. Moreover, to enhance predictive capabilities further, we integrate these individual methodologies through ensemble modeling, resulting in our ensemble model exhibiting a remarkable 96% accuracy in the early detection of CKD. This significant advancement holds immense promise for improving clinical outcomes and underscores the critical role of machine learning in addressing complex medical challenges.

## I. INTRODUCTION

According to the 2010 Global Burden of Disease study, chronic renal disease jumped from 27th place in 1990 to 18th place in 2010 among all causes of death worldwide. 10% of the global population is affected by Chronic kidney disease (CKD), and due to a lack of awareness and affordable treatment, millions of people die from it every year. A disease is called when an abnormal condition of an organism interrupts the normal bodily functions that often lead to feelings of pain and weakness, and is usually associated with symptoms and signs that lead to dysfunction, distress, or even death[1]. The kidneys are one of the main organs of the human body. To have a healthy life, healthy kidneys are a must. The kidneys remove wastes and extra water to make urine and filter about a half cup of blood in every minute in the human body[2]. Furthermore, the kidneys also maintain a balance of water, salt, and minerals by removing acid from the body[3]. When the kidneys do not function properly to filter blood into the body that is a situation called kidney disease. Chronic kidney disease (CKD) involves a gradual loss of kidney function that leads to the accumulation of excess waste and fluids in our body[4]. The waste that has accumulated in the body can be harmful to our overall health leading to end-stage renal disease when the kidneys stop working completely[5].

Ninety percent of the 37 million persons with CKD in the US are unaware that they have the disease. Many CKD patients do not exhibit any symptoms until their condition reaches more advanced stages or until complications arise[6]. If symptoms do appear, they could consist of Urine with foam, Spitting, Urinating (more or less frequently than usual), Dry or itchy skin, Tiredness and nausea, Appetite decline, and Loss of weight without attempting to reduce it. Individuals with more severe stages of CKD could moreover observe: Difficulty focusing, Tingling or edema in their legs, Breathing difficulties, Vomiting, Difficulty falling asleep, and Ammonia, also referred to as "fishy" or Urine-like, smelled in breath[7].

There are many causes of CKD, but these two are the most frequent causes of i) diabetes and ii) hypertension; by diabetes approximately 25% of cases of kidney failure are caused, while the other 33% are caused by hypertension. High blood pressure is one of the major reasons for CKD[7]. If someone has long-term conditions like hypertension, diabetes, or high blood pressure that can lead to life-threatening CKD[8]. Smoking can cause many cardiovascular diseases that can often lead to CKD. Quitting smoking, stopping drinking alcohol, eating healthy, exercising regularly, and being careful about taking painkillers can prevent CKD from happening[9]. CKD is a worldwide public health emergency. According to the World Health Organization, there were approximately 58 million deaths globally in 2005, with 35 million related to chronic disease[10]. Unfortunately, only 2 million people get dialysis or a kidney transplant for their survival; however, this number represents only 10% of those who require treatment[11].

To assess someone's kidneys and make the diagnosis of kidney failure, a medical professional may employ a range of renal function tests such as i)Blood test ii) Urine test, and iii) Imaging tests. Based on someone's estimated glomerular filtration presence rate in the body, there are different stages of kidney disease (eGFR) such as 1) Stage I(eGFR is higher than 90 but below 100) 2) Stage II (eGFR is higher than 60 but below 89)3) Stage III(eGFR is higher than 30 but below 59) 4)Stage IV(eGFR is higher than 15 but below 29) 5)Stage V(eGFR is below 15) [12]. Medical officials also give blood tests to discover kidney problems. If the kidneys are producing urine along with protein and blood then the kidneys are not working perfectly[13]. Then there can be two possible scenarios to solve kidney problems, one is dialysis and another one is kidney transplant[14]. Dialysis is a very physically and mentally challenging process for the patient, and the medical official must be concerned about the patient's food habits, age, gender, body movements, and how often the dialysis is being taken[15]. When the dialysis fails to improve the patient, then there is only one option left for the doctors, and that is a kidney transplant. It is a very costly and challenging process and also requires the finest medical officials[16].

In the era of Artificial Intelligence(AI) and Machine Learning(ML), researchers are coming up with many ideas to detect CKD not only using clinical diagnosis but also with the help of various algorithms and models[17]. To detect CKD medical images and physiological signals are being used by researchers in the Deep Learning (DL) techniques[18]. Various models and studies have been conducted and found that most of the researchers used the Conventional Neural Network(CNN) model, which did not perform well on multi-class image classification[19].

After studying many researches we are going to propose a model for the early detection of CKD with more accuracy. So in this study, we are trying to leverage the transfer learning models and custom Convolutional neural network(CNN) to classify CKD. We have trained 3 transfer learning models and among them, MobileNetV2 performed well with more than 90% accuracy. Here in this research, we have used state-of-the-art transfer learning models i.e., i) EfficientNetV2 ii)InceptionNetV2, and iii)MobileNetV2. We have also used the Vision Transformer (ViT) method for training our model, and we have got 91.5% accuracy. Finally, we proposed a model that used ensemble methods and we achieved 96% accuracy. We are proposing an ensemble model for the early detection of CKD so that much suffering can end and lives can be saved.

II. LITERATURE REVIEW

This section reviews some recent image-processing research works. The paper discusses several methodologies and strategies for using image processing to diagnose Kidney disease.

Singh et al. [20] proposed a framework for early prediction of chronic kidney disease using a deep neural network. The authors of this study examined the Recursive Feature Elimination technique to determine which features are crucial for making accurate predictions. For the aim of classification, they fed multiple features to machine learning models. The proposed Deep neural model achieved better than the SVM, KNN, Logistic regression, Random Forest, and Naive Bayes algorithms. The suggested model's main flaw was that it could only be validated using limited sample sizes of data. In order to better recognize the severity of CKD, large amounts of progressively high-quality and representative CKD data will be gathered in the future.

Majid et al. [21] conducted transfer learning strategies for kidney disease classification using CT images. In order to enhance the efficiency of the training procedure, the researchers used a range of pre-processing approaches and employed imagine scaling methods. Within this framework, they unveiled two improved Transfer Learning (TL) models, ResNet-101 and DenseNet-121, for predicting kidney tumors. DenseNet-121, a Tranfer learning model, performed at an incredible 98.22% accuracy. The study's comparative analysis, model testing, and generalizability exhibited some drawbacks. The comparative analysis of detection procedures, such as Random Forest, Support Vector Machine, Gradient Boosting, Light Gradient Boosting Model, and deep learning models ResNet101 and DenseNet-121, has been limited in its scope, missing to fully explore other pertinent models.

Sudharson et al. [22] applied an ensemble of deep neural networks using transfer learning for kidney ultrasound image classification. The pre-trained DNN models are applied to three different datasets for feature extraction, and a support vector machine is then used for classification. The process involves the combination of several pre-trained deep neural networks, including ResNet-101, ShuffleNet, and MobileNet-v2. The final predictions are made by the use of the majority voting approach. The approach that was proposed achieved a maximum classification accuracy of 96.54% when tested with quality images and 95.58% when tested with noisy images.

Kim et al. [23] applied an artificial neural network for chronic kidney disease classification. The GLCM technique, extensively utilized in ultrasound image processing, was used to extract parameters from each ROI. The artificial neural network (ANN) has 58 input parameters, ten hidden layers, and three output layers. The concluded classification rate was 95.4% using the ANN model, and the training epoch was 38 times. They will apply the Transfer learning model to increase performance on this dataset and also will increase the dataset for the training model.

Radya et al. [24] applied data mining techniques for kidney disease prediction. These techniques included Multilayer Perceptrons, Support Vector Machines, Radial Basis Functions, and Probabilistic Neural Networks. The PNNs algorithm has the greatest overall classification accuracy percentage of 96.7% when compared to other algorithms for identifying the stages of patients with chronic kidney disease (CKD). Using four distinct algorithms, The authors used very limited datasets, consisting of just 361 instances, to forecast chronic kidney disease (CKD).

Bhandari et al. [25] proposed a lightweight convolutional neural network to detect kidney disease. The training data's means and standard deviation were used by The LIME image explanation to generate a number of features and changes. LIME gave a visual description of how the model made its decisions and pointed out the parts of the image that were most important for predicting a specific category. After extensive testing, the suggested CNN model proved to be nearly perfect, with an accuracy of 98.68 percent. This study only employed a small number of CT scans. Therefore, the results may be better if the use of data augmentation. By integrating DL models with other XAI algorithms, they will increase the clarity of the outcomes.

Bhattacharjee et al. [26] proposed a computer-aided diagnostic model for kidney disease classification using a modified Xception deep neural network version, with image net weights derived via transfer learning. The model trained with these two datasets has a 99.39% success rate in this research. Due to the ensemble models' lack of depth, it is impossible to extract contextual information from adjacent slices. A 3D classification model that makes use of interc-slice context is one potential solution to this issue.

Kanwal et al. [27] proposed an automated model for the classification of kidney abnormalities. Two useful automatic models were included in the proposed study. First was Efficient-b0, and second was ResNet-18. Both of them correctly predicted problems with the kidneys. The accuracy of the proposed ResNet-18 model was determined to be 98.1% after testing. They will gather real-time datasets in order to test and train their model as part of the next phase of their research.

Wasi et al. [28] proposed an identification model for kidney tumors using transfer learning. In order to identify kidney tumors from CT scans, a deep convolutional neural network (DCNN) based transfer learning approach is proposed. Results for the 5,284-image test set showed an accuracy of 92.54% after 50 epochs of training. They only used data, including images of kidneys or tumors. For further extension activities include enhancing the quantity and quality of datasets, which includes more images of kidneys, tumors, implants, and other abnormalities that may impact the model's ability to identify kidney malignancies accurately.

Kadir Yildirim et al. [29] proposed an automated system for detecting kidney stones in the human body using coronal computed tomography (CT) images with the help of an Artificial Intelligence technique, namely Deep Learning. About 1800 images were used for each person's cross-sectional CT images. This system can detect small-size kidney stones with an accuracy of 96.82%. This method can be used in urology to solve many problems for clinical application because it has given great results for a larger dataset of 433 subjects.

Fuzhe Ma et al. [30] stated that chronic kidney disease (CKD) is increasing day by day, to diagnose CKD, machine learning techniques have become an essential tool in recent years. They suggested a model using a Heterogeneous Modified Artificial Neural Network (HMANN) to detect, segment, and diagnose CKD on the platform of the Internet of Medical Things (IoMT) that is described as a Support Vector Machine (SVM) and Multilayer Perceptron (MLP) using a Backpropagation (BP) method. During the preprocessing step, ultrasound images are used to segment the image. The proposed algorithm reduces the time and provides an accuracy of 97.5 percent.

Nicholas Heller et al. [31] reported that numerous studies have been conducted to establish a connection between the geometric and anatomical features of kidney tumors and the outcomes of oncology. Producing high-quality 3D segmentations takes a lot of time and human energy from the tumors and the kidneys that host them. Furthermore, in autonomous 3D segmentation, deep learning techniques have achieved good results, and they need a tremendous amount of training data. In 2019, the International Conference on Medical Image Computing and Computer-Assisted Intervention (MICCAI) organized the Kidney and Kidney Tumor Segmentation Challenge (KiTS19) to encourage innovations in the automatic segmentation problem. In this study, 90 cases were predicted based on the average Sørensen-Dice coefficient between the kidney and tumor. The Winning team set a benchmark for 3D semantic segmentation with an accuracy of 97.4% for the kidney and 85.1% for the tumor.

Swapnita Srivastava et al. [32] proposed a model to determine and diagnose Chronic Kidney Disease using computational-based methods. This study depends on data on chronic renal disease available on the public platform. In this proposed model, a performance-tuning nested approach is used that takes into account adjusting hyper-parameters as well as determining the appropriate weights to join ensembles (Ranking Weighted Ensemble). The result of the study gives an accuracy of 98.75%, and it could be used to develop an automated system that can detect kidney disease.

Navaneeth Bhaskar et al. [33] suggested a new model to find out kidney disease automatically using machine learning algorithms. To identify the disease, the salivary urea concentration is observed with a new sensing approach to monitor the urea levels in the saliva sample. A 1-dimensional (1-D) deep learning Convolutional Neural Network (CNN) method that includes a Support Vector Machine (SVM) classifier this study has developed. The accuracy of the model has been enhanced because of the CNN-SVM integrated network. The proposed model shows that it provides 98.04% accuracy.

Francesco Paolo Schena et al. [34] proposed an artificial neural network prediction model for end-stage kidney disease (ESKD) in patients with primary immunoglobulin A nephropathy (IgAN). To predict ESKD, this study applies a two-step procedure of a classifier model and to detect the development of ESKD, a regression model is applied. A clinical decision support system (CDSS), which is easy to use, has been developed to predict ESKD in patients with IgAN with a median follow-up of 5 and 10 years. The accuracy of the classifier model is 89% for the patients with a follow-up for ten years. The proposed system gives a result of 91% to predict IgAN in a patient.

Guozhen Chen et al. [35] proposed a method to detect Chronic Kidney Disease (CKD) in the early stage efficiently and effectively by using various deep learning methods, and Adaptive hybridized Deep Convolutional Neural Networks (AHDCNN). This study stated that to achieve high accuracy, an algorithm model has been developed using a Conventional Neural Network (CNN) to classify the dataset properly applying feature dimension. The Internet of Medical Things platform (IoMT) concluded that using machine learning techniques helped to produce the solution to kidney disease as well as other diseases too. The proposed system can achieve an accuracy of 97.3% in detecting CKD.

Md Nazmul Islam et al. [36], to diagnose kidney disease properly at the earliest time, an AI-based system needs to be developed. In this study, three main renal disease categories, kidney stones, cysts, and tumors, are discussed to develop an AI-based kidney disease diagnostic system using a total of 12,446 CT whole abdomen and urogram images. After analyzing the data, the study found that the images had the same type of mean color distribution from all of the classes. To find the best result, six machine learning models were used, namely EANet, CCT, and Swin transformers, Resnet, VGG16, and Inception v3. The results show that the VGG16 and CCT models provide a decent output in terms of accuracy, but the swin transformer gives an accuracy of 99.30%.

Chin-Chi Kuo et al. [37] proposed a deep learning approach for automatically determining the estimated glomerular filtration rate (eGFR) and Chronic Kidney Disease (CKD) status. In this study, to predict kidney function, they used 4,505 kidney ultrasound images labeled using eGFRs. This study also revealed that a neural network architecture is used for the transfer learning technique along with the ResNet model on an ImageNet dataset. Furthermore, this study used kidney length annotations to remove the peripheral region of the kidneys and applied various data augmentation schemes to produce additional data with variations to extract more information from the ultrasound images. Bootstrap aggregation was applied to improve the model's performance and avoid overfitting. The proposed AI-GFR model provides an accuracy of 85.6%, which shows that this model can be applied to detect CKD status in clinical practice.

III. METHODOLOGY

A. Dataset

*1) Data Collection:* The dataset, aptly named "CT KIDNEY DATASET: Normal-Cyst-Tumor and Stone" [38], was meticulously assembled from diverse medical sources, specifically various hospitals in Dhaka, Bangladesh. Patients within this dataset had previously received diagnoses related to kidney conditions, covering an extensive array of scenarios. The dataset is notably comprehensive, featuring 12,446 unique instances. These cases include 3,709 instances of cysts, 5,077 normal cases, 1,377 instances of stones, and 2,283 tumor cases. Notably, both contrast and non-contrast studies, as well as Coronal and axial cuts, contribute to the dataset's richness and representativeness of diverse kidney pathologies.

B. Data Pre-processing

*1) Image Augmentation:* A pivotal aspect of this research involves the strategic application of data augmentation techniques, realized through the utilization of the 'ImageDataGenerator' class. The augmentation strategy encompasses a range of transformations, including rescaling (224x224), rotation, zooming, horizontal and vertical flipping, and shifting. These augmentations serve the dual purpose of diversifying the dataset, enriching it with varied instances, and enhancing the model's resilience to the inherent variations in kidney images.

*2) Train-Test Split:* To meticulously assess the generalization capabilities of the models, an 80-20 train-test split was implemented. This careful partitioning ensures that the models are trained on a substantial dataset while retaining a sufficiently independent test set for rigorous evaluation. The adoption of an 80-20 split aims to strike a balance between model convergence and the prevention of overfitting.

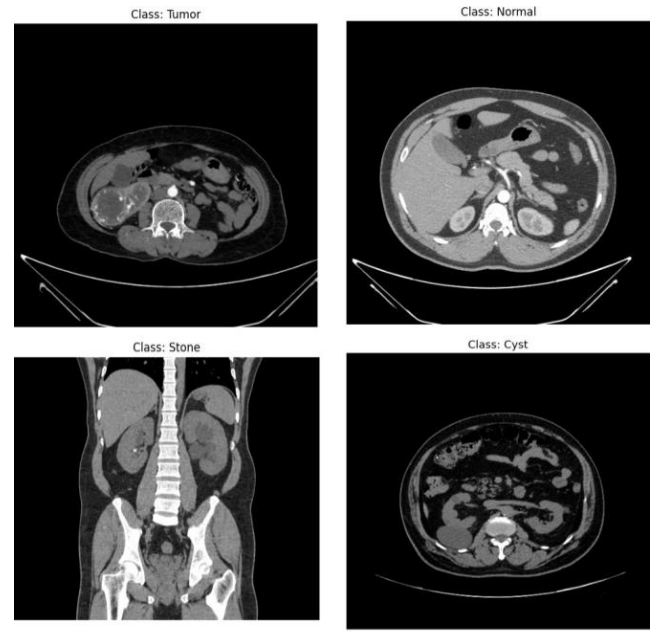

Fig. 1. Random Images From Dataset

*3) Label Encoding:* The process of assigning numeric labels to distinct classes was carried out through the application of the 'LabelEncoder' from the scikit-learn library. Following this, a crucial step involved the conversion of these labels into one-hot encoding. This categorical representation is fundamental for training the models, enabling them to accurately discern and classify the diverse array of kidney conditions present in the dataset.

*4) Dataset Creation:* The creation of datasets for both the training and validation phases was meticulously executed through the 'flow_from_dataframe' method from TensorFlow. The training dataset was intentionally enriched with augmentations, thereby exposing the model to an even broader array of representations of kidney conditions. In contrast, the validation dataset remained unaltered. ensuring that the model's performance could be evaluated on authentic, real-world, and unaltered data.

C. Transfer Learning Models

*1) EfficientNetV2:* EfficientNetV2 [39], a family of convolutional neural network (CNN) models, introduces a novel approach known as compound scaling to systematically enhance the model's depth, width, and resolution in tandem, resulting in improved overall performance. This architecture features a sequence of mobile inverted bottleneck blocks similar to those found in MobileNetV2 but with increased depth and width. Through intelligent employment of compound scaling, EfficientNetV2 achieves a fine balance between accuracy and computational efficiency. In our research, we leverage EfficientNetV2 as a state-of-the-art model for image classification tasks, utilizing it as a benchmark to evaluate its classification

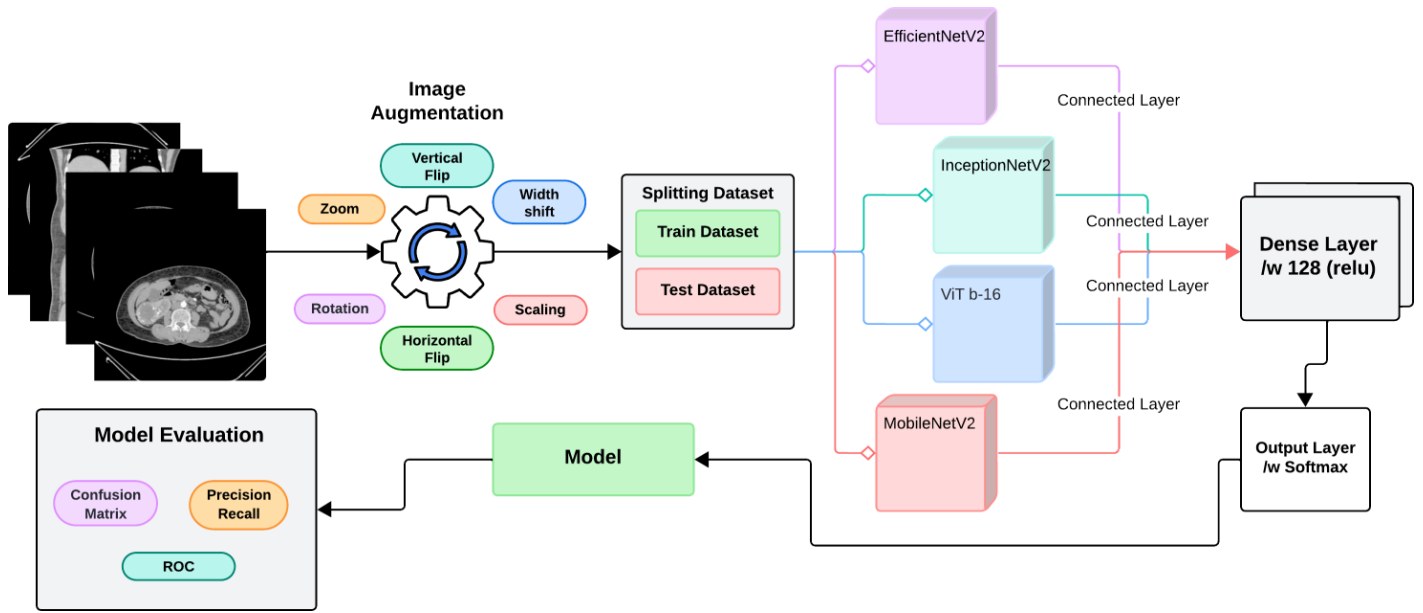

Fig. 2. Methodology of Proposed Architecture

capabilities and to compare it against other contemporary architectural paradigms.

*2) InceptionNetV2:* InceptionNetV2 [40] stands as a noteworthy family of convolutional neural network (CNN) models, extending the original InceptionNet with innovative architectural enhancements. InceptionNetV2 captures features at different resolutions thanks to its unique inception modules that use multi-scale convolutional filters. The model further integrates factorized convolutions to manage computational complexity without compromising representational power. Unlike EfficientNetV2's compound scaling, InceptionNetV2 relies on meticulously designed modules and architectural refinements to strike a balance between model intricacy and performance. Featuring sophisticated blocks, including Inception and reduction blocks, InceptionNetV2 excels in effective feature extraction. Our comparison study uses InceptionNetV2 as a key part because it lets us compare its image classification performance to other modern, state-of-the-art models. This shows how model design, complexity, and classification accuracy are all connected in the world of image recognition tasks.

*3) MobileNetV2:* MobileNetV2 [41] architecture caters to mobile and embedded devices characterized by limited computational resources. It strategically decomposes the conventional convolution operation into depthwise and pointwise convolutions, effectively reducing computational overhead. Notably, MobileNetV2 employs depth-wise separable convolutions, a key innovation contributing to its efficiency. Through the utilization of inverted residual blocks with linear bottleneck layers, the architecture strikes an optimal balance between model compactness and accuracy. These blocks facilitate effective feature extraction while simultaneously minimizing the parameter count. In our study, MobileNetV2 serves as a comparative benchmark, enabling us to assess its performance relative to other contemporary state-of-the-art models.

*4) Vision Transformer:* The Vision Transformer (ViT) represents a paradigm shift in image classification, moving away from traditional convolutional neural networks (CNNs) towards transformer-based architectures. Pioneered by Dosovitskiy et al. [42], ViT applies self-attention mechanisms and transformer architecture originally developed for natural language processing tasks to image recognition. By breaking down input images into fixed-size patches, ViT facilitates direct processing through transformer layers, capturing global context information effectively. Despite the lack of hierarchical feature extraction inherent in CNNs, ViT compensates with its ability to learn long-range dependencies through self-attention, enabling robust feature representation. In our research, ViT serves as a crucial benchmark for evaluating its classification capabilities alongside other contemporary state-of-the-art models. The goal of this comparison study is to show how transformer-based architectures, computational efficiency, and classification accuracy all work together in complex ways. This will help the field of image recognition research as it grows.

*D. Model Configuration*

The training of each transfer learning model was conducted with meticulous care over 50 epochs. This iterative process sought a harmonious convergence of the model while guarding against the potential pitfalls of overfitting. The training was executed leveraging the augmented training dataset, with subsequent validation on an unaltered validation dataset. Softmax as the activation function and Adam as the optimizer were used. Evaluation metrics, including categorical cross-entropy loss and accuracy, were employed to critically assess model performance. The early stopping feature, along with the patience of 5 epochs, was also very important for improving the efficiency of training and reducing the chance of overfitting, which in turn made the models more robust.

*E. Ensemble model*

The ensemble model is an ensemble architecture that amalgamates multiple state-of-the-art image classification models,

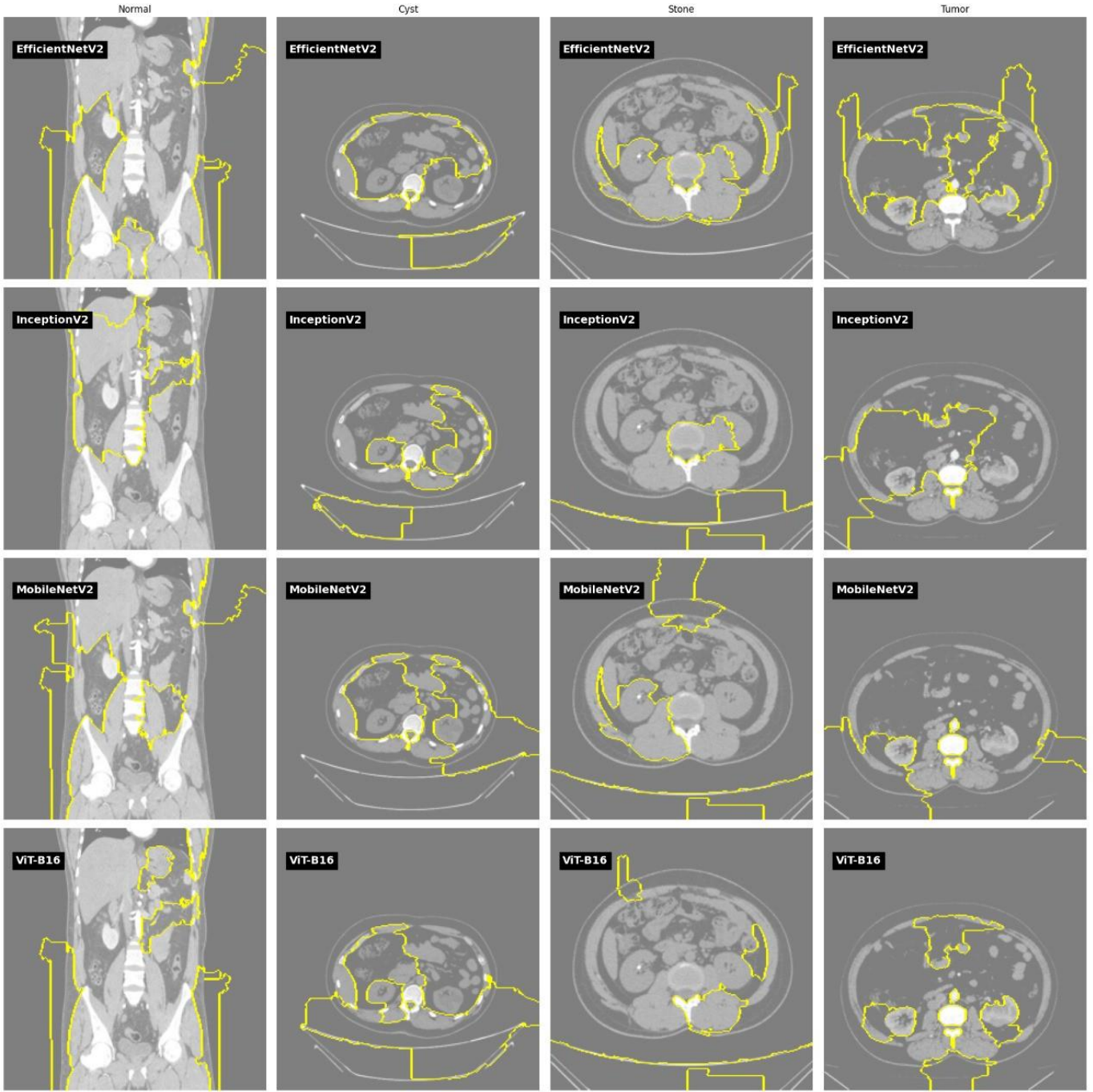

Fig. 3. Methodology of Proposed Architecture

including convolutional neural networks (CNNs) and the Vision Transformer (ViT), in a harmonious manner. This ensemble approach is a strategic endeavor to harness the distinct strengths of diverse architectural paradigms and capture multifaceted feature representations, thereby potentially enhancing the overall classification performance. The model's design is a symphony of parallel branches, each composed of a pre-trained model architecture, that converges through a concatenation layer, allowing for the seamless fusion of hierarchical and global representations. The subsequent dense layers and the output layer orchestrate the amalgamated features, culminating in a robust ensemble model poised to deliver superior classification accuracy.

### F. Explainable AI (XAI) Using LIME

In order to improve the comprehensibility of our deep learning model for detecting kidney disease, we utilized Explainable AI (XAI) methods, specifically employing Local Interpretable Model-agnostic Explanations (LIME). LIME offers clear and explicit explanations for the predictions provided by complicated models. It achieves this by generating simpler and more interpretable models that approximate the behavior of the 11 complex models for specific cases. This study employed LIME to detect and emphasize areas in medical images that had a significant effect on the decision-making process of the model. This provided crucial insights into the primary features that drove the model's predictions. Figure 3 illustrates the visual outcomes of implementing LIME on four different deep learning models: InceptionNet-V2, MobileNet-V2, EfficientNetV2, and Vision Transformer (ViT-B16). Each model was assigned the duty of classifying images under four different categories: Normal, Cyst, Stone, and Tumor. The yellow highlighted por tions in the figure indicate the areas of the images that had the most influence on the model's predictions. After implementing LIME, it was seen that certain models had exceptional performance for specific classes, as depicted in the figure. EfficientNetV2 showed a robust capability in accurately detecting areas that indicated the presence of cysts and tumors, whereas MobileNet-V2 proved to be more efficient in classifying normal and stone classes. In order to enhance the overall performance in terms of both robustness and accuracy, we combined these four models into an ensemble, focusing on their distinct advantages across various classes. The use of this ensemble technique guaranteed that the best possible model was not only precise but also dependable for all classifications of kidney disease. The incorporation of XAI using LIME pro vided a lucid comprehension of the decision-making process employed by the models, hence bolstering the reliability and interpretability of our ensemble model in clinical environments.

## IV. PERFORMANCE ANALYSIS

### A. Performance of Transfer Learning Models

The research studies applied pre-trained InceptionV3, EfficientNet, and MobileNet models and also applied ViT to determine the most significantly effective model for identifying and classifying Kidney disease using the CT kidney dataset. Our models were designed utilizing 128x128 images as input data. To fine-tune the hyperparameters, a batch size of 32 and 50 epochs was employed during the training phase. Given the multi-class nature of the dataset, the softmax activation function was applied in the output layer. The model was compiled using the Adam optimizer and the categorical_crossentropy loss function. We achieved the ability to obtain an average accuracy for three different models, such as MobileNet-V2, achieving 87.25% accuracy, EfficientNet-V2, achieving 86.75% accuracy, InceptionNet-V2, achieving 83.25% accuracy, and ViT, achieving 91.5% accuracy.

Five performance indicators - accuracy (ACC), precision (PPR), recall or sensitivity (Sen), F1 score, and Area under the ROC curve (AUC) score - have been adapted for use with every analysis dataset in order to evaluate the proposed kidney disease classifier. On average, the mobileNet-V2 model obtained a score of 87.25% in terms of precision, 87.5% in terms of recall, 87.25% in terms of f1-score, and 92% in terms of AUC Score. For the efficientNet-V2 model, the average score obtained was 86.75% accuracy, 85.5% recall, 88.5% f1-score, and 90.75% AUC Score. For the InceptionNet-V2 model, the average scores were 83.25% precision, 79.75% recall, 80.75% f1-score, and 87.25% AUC score. For Transformer ViT, the average scores were 91.5% precision, 89.25% recall, 90% f1-score, and 93.25% AUC score. The performance indicators for each class are briefly presented in Table I for all models.

TABLE I. PERFORMANCE ANALYSIS OF DIFFERENT MODELS

| Model (Class) | Precision | Recall | F1-Score | AUC |
|---|---|---|---|---|
| **MobileNET-V2** | | | | |
| Tumor | 0.87 | 0.98 | 0.92 | 0.96 |
| Cyst | 0.96 | 0.89 | 0.92 | 0.93 |
| Normal | 0.79 | 0.80 | 0.80 | 0.89 |
| Stone | 0.87 | 0.83 | 0.85 | 0.90 |
| **EfficientNet-V2** | | | | |
| Tumor | 0.94 | 0.89 | 0.91 | 0.93 |
| Cyst | 0.90 | 0.95 | 0.93 | 0.94 |
| Normal | 0.82 | 0.72 | 0.87 | 0.85 |
| Stone | 0.81 | 0.86 | 0.83 | 0.91 |
| **InceptionNet-V2** | | | | |
| Tumor | 0.90 | 0.91 | 0.90 | 0.93 |
| Cyst | 0.86 | 0.91 | 0.89 | 0.91 |
| Normal | 0.77 | 0.54 | 0.63 | 0.76 |
| Stone | 0.80 | 0.83 | 0.81 | 0.89 |
| **ViT** | | | | |
| Tumor | 0.90 | 0.98 | 0.94 | 0.97 |
| Cyst | 0.95 | 0.95 | 0.95 | 0.96 |
| Normal | 0.87 | 0.82 | 0.84 | 0.90 |
| Stone | 0.94 | 0.82 | 0.87 | 0.90 |

The confusion matrix for attentiveness models, namely mobileNet-V2, EfficientNet-V2, InceptionNet-V2, and ViT, is shown in Figure 3, respectively. The more efficient performance of the three deep learning models is evident when applied to CT Kidney datasets, with the method of attention of these models proving to be particularly effective. It is apparent that the classifier successfully classified a significant portion of the instances.

The ROC Curve plots for three transformed deep learning architectures, namely mobileNet-V2, EfficientNet-V2, InceptionNet-V2, and ViT, are shown in Fig. 4, respectively. The models effectively distinguished between all positive and negative classes with a high degree of accuracy, as shown by the considerable AUC for all anomalies. Based on the test data, the ViT model predicted a very high true-positive rate(TPR), achieving 97% of the AUC values for tumor detection, which was important since the ROC curve was reliant on the TPR and the FPR and average AUC score is 93.25%. MobileNetV2 model also performed the highest average score from transfer learning models. Therefore, even for classes with non-uniform sample distributions, these findings indicated that the Transformer ViT model was more robust and consistent.

Precision is a measure of accuracy that evaluates the proportion of accurately identified positive samples (True Positive) out of the total number of identified positive samples. It serves to analyze the validity of the machine learning model in classifying the model as positive. The percentage of positive samples that are accurately identified as positive samples to the total number of accurate positive samples is called recall. The effectiveness of the model in identifying positive samples can be measured by the recall. When it comes to producing a great machine learning model that generates outcomes that are more precise and accurate, these values are essential. Following the completion of the measurement, we were able to achieve the result shown in Figure 5 for precision compared to recall. When compared to other transfer learning models, ViT obtained a much higher score, which had an average score of 96.75%, than other models. ViT model achieved significant precision-recall scores for tumors and cysts, 99% and 99%, respectively.

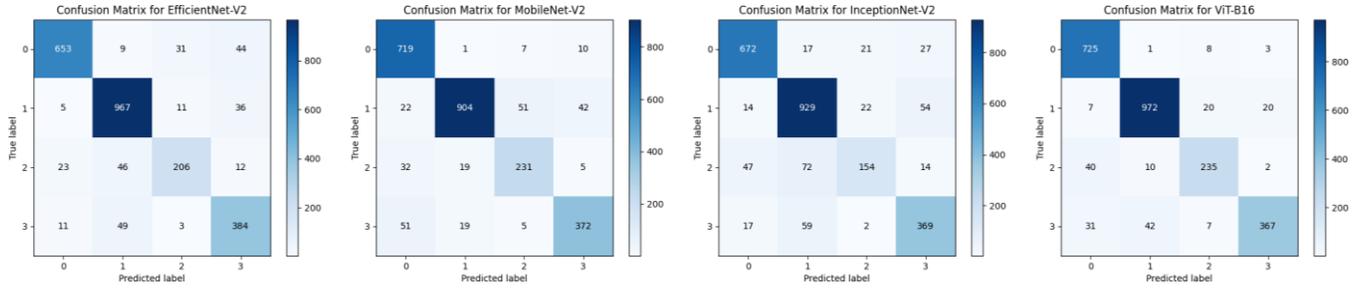

Fig. 4. Confusion matrix for four models

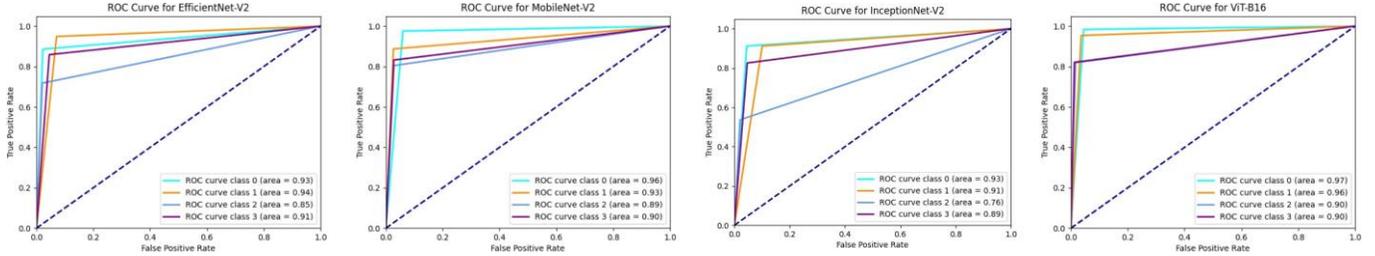

Fig. 5. ROC Curve for four models

After analyzing all models, we show that the ViT model achieved the highest score for precision, recall, f1-score, AUC score, and precision-recall score. When compared to all transfer learning models, the mobileNetV2 model also performed significantly in detecting Kidney disease identification.

### B. Performance of Ensemble Model

We used the ensemble method for the proposed work, which actually based on ViT and a pre-trained model. After training our ensemble model, we get outperforming results that accurately predict kidney disease. We achieved significant results from the ensemble model accuracy is 96%, recall and f1-score are 97% and 96.5%, respectively. When we compare our ensemble model performance from other baseline models where we got precision value for the tumor is 98%, the cyst is 100%, 92%, and 94%, which is a satisfactory score than baseline MobileNet-v2, EfficientNet-V2, InceptionNet-V2, and ViT models which we see to Table-I. We show in Table II that other performance indicators recall, f1-score, and AUC score, are impressive for our ensemble model than baseline models to compare Table I.

In comparison to the curve, the ensemble model Performed significantly better for individual's ROC and Precision-recall curve than other distinct models. The ensemble model achieved a 98% average score from the ROC curve and a 99.5% score from the precision-recall score. We decided that after all performances, the ensemble model performed superior and robustly, which will contribute identification of CKD in the medical sector.

### C. Comparison with Existing Models

The top-performing models from both the most recent literature and the proposed study are compared and analyzed based on their evaluation accuracy, as shown in Table III. The findings show that in order to move the field forward and become better at identifying images and related tasks, it's crucial to look at and compare various models and approaches. The ensemble model's most incredible accuracy of 90% and AUC of 95% demonstrated the model's effectiveness in assisting clinical decision-making about the prognosis of kidney disease.

TABLE II. PERFORMANCE ANALYSIS OF THE ENSEMBLE MODEL

| Classes | Precision | Recall | F1-Score | AUC |
|---|---|---|---|---|
| Tumor | 0.98 | 0.96 | 0.97 | 0.97 |
| Cyst | 1.0 | 0.98 | 0.99 | 0.99 |
| Normal | 0.92 | 0.94 | 0.93 | 0.97 |
| Stone | 0.94 | 1.0 | 0.97 | 0.99 |

TABLE III. COMPARISON WITH EXISTING WORKS

| References | Dataset Size | Model(s) | Accuracy |
|---|---|---|---|
| Zabihollahy et al. [43] | 315 | Convolutional neural network | 83.75% |
| Akgun et al. [44] | 460 | MobileNet | 86.42% |
| | | ResNet50 | 82.06% |
| Kalkan et al. [45] | 5000 | ResNet152V2 | 89.58% |
| | | MobileNetV2 | 88.80% |
| Lee et al. [46] | 1596 | Inception V3 | 74.3% |
| | | MobileNet | 72.37% |
| This Work | 12466 | Ensemble Method | 96% |

### V. CONCLUSION

Early detection and classification of kidney disease may save human lives. Detection procedures that are done manually are often laborious and dependent on the knowledge of medical professionals. The advancement of automated classification systems is thus highly encouraging, as they provide reliable and quick outcomes. Improved accuracy and reliability in kidney disease detection may be achieved by utilizing deep learning methods such as pre-trained models and ViT. According to our findings, ViT had the highest accuracy (90%) when compared to all of the fine-tuned models, then MobileNet-V2, EffecientNet-V2, and InceptionNet-V3. However, The ensemble model achieved a robust performance that is 96% which is more accurate than baseline models. The research also conducted a manual error analysis to improve the performance of the pre-trained models. This might result in more accurate diagnoses and improved treatment choices for individuals with kidney diseases.

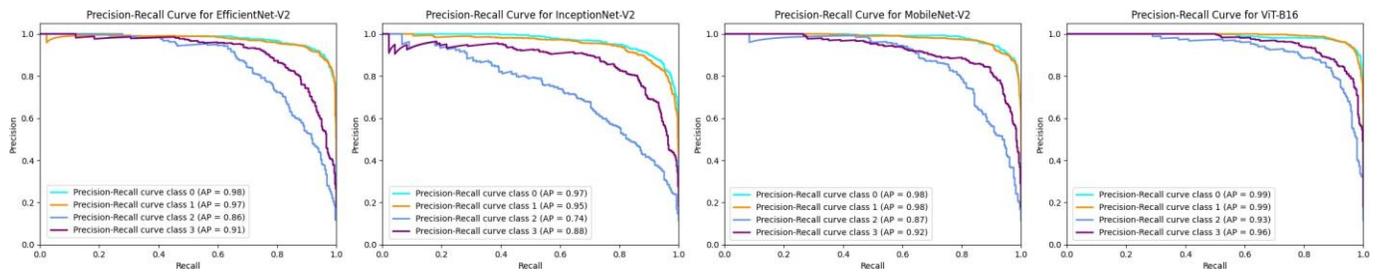

Fig. 6. Precision vs Recall for four models

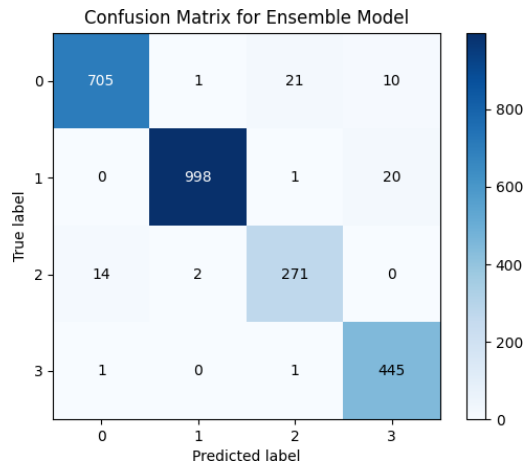

Fig. 7. Ensemble Model Confusion Matrix

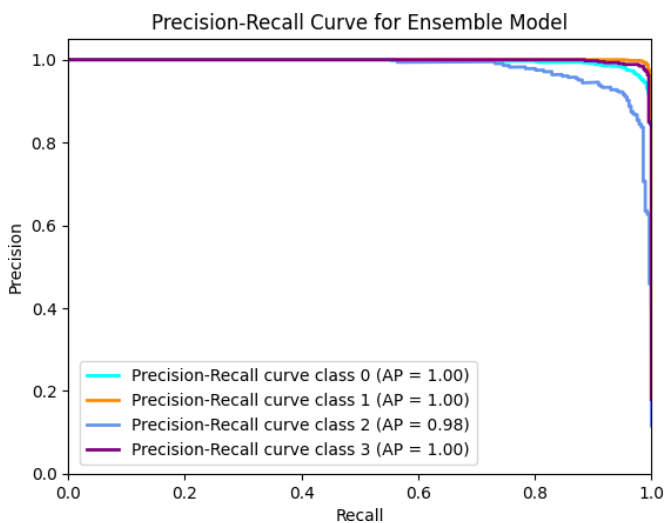

Fig. 8. Ensemble Model Precision Recall

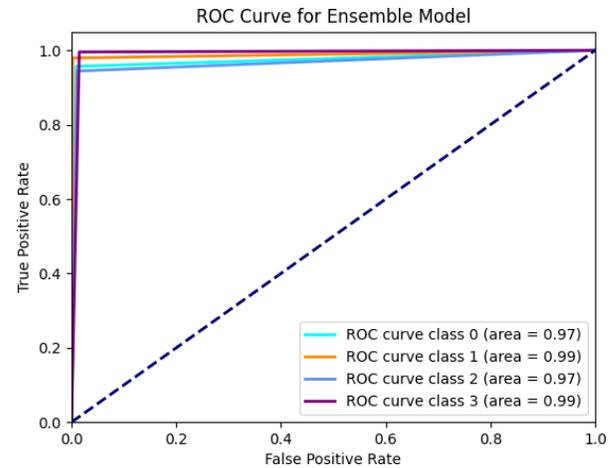

Fig. 9. Ensemble Model ROC

For future studies, it would be prudent to experiment with the MobileNet-V2, EffecientNet-V2, and InceptionNet-V2 optimized Tranfer learning models and ensemble models on other datasets to see how well they hold up in different types of clinical situations. To make the proposed strategy more practical, it is essential to study other datasets with diverse demographics, patient groups, and quality imaging levels. An in-depth review of the model's abilities and shortcomings will reveal any adjustments that may be necessary for its successful implementation in clinical practice. Finally, by tackling these potential future research directions, we may create a kidney tumor identification approach that is more comprehensive and flexible, making it applicable and successful in more clinical circumstances.

## DECLARATIONS

- Funding: No funding was received.

- Conflict of interest: The authors declare that they have no conflict of interest.

- Ethics approval and consent to participate: Not applicable. The study uses open-source data that does not involve human participants directly.

- Data Availability: The data supporting this study's findings is available online. Available: https://www.kaggle.com/datasets/nazmul0087/ct-kidney-dataset-normal-cyst-tumor-and-stone

- Consent for publication: Not applicable.

- Code availability: The Code supporting the findings of this study will be made available upon request from the authors.

- Author contribution: All authors contributed to the study conception and design. Material preparation, data collection, and analysis were performed by Iftekhar Ahmed, Md. Jalal Uddin Chowdhury and Shadman Sakib. The first draft of the manuscript was written by all authors who commented on previous versions of the manuscript. All authors read and approved the final manuscript.

## REFERENCES

[1] B. Online, "Disease definition and examples - biology online dictionary," February 2022. [Online]. Available: https://www.biologyonline.com/dictionary/disease

[2] C. Clinic, "Kidney," accessed on April 29, 2024. [Online]. Available: https://my.clevelandclinic.org/health/body/21824-kidney

[3] National Institute of Diabetes and Digestive and Kidney Diseases, "Your kidneys & how they work," October 2023. [Online]. Available: https://www.niddk.nih.gov/health-information/kidney-disease/kidneys-how-they-work

[4] Mayo Clinic, "Chronic kidney disease - symptoms and causes," September 2023. [Online]. Available: https://www.mayoclinic.org/diseases-conditions/chronic-kidney-disease/symptoms-causes/syc-20354521

[5] American Kidney Fund, "Types of kidney diseases," November 2023. [Online]. Available: https://www.kidneyfund.org/all-about-kidneys/types-kidney-diseases

[6] Centers for Disease Control and Prevention, "Chronic kidney disease basics," accessed on April 29, 2024. [Online]. Available: https://www.cdc.gov/kidneydisease/basics.html

[7] National Kidney Foundation, "Facts about chronic kidney disease," December 2023. [Online]. Available: https://www.kidney.org/atoz/content/about-chronic-kidney-diseasecauses

[8] National Institute of Diabetes and Digestive and Kidney Diseases, "High blood pressure & kidney disease," October 2023. [Online]. Available: https://www.niddk.nih.gov/health-information/kidney-disease/high-blood-pressure

[9] NCD Alliance, "Chronic kidney disease," March 2022. [Online]. Available: https://ncdalliance.org/why-ncds/ncds/chronic-kidney-disease

[10] National Kidney Foundation, "Global facts: About kidney disease," November 2023. [Online]. Available: https://www.kidney.org/kidneydisease/global-facts-about-kidney-disease

[11] R. Mathur, "How home healthcare can help patients with serious kidney ailments keep their condition under check," March 2022. [Online]. Available: https://timesofindia.indiatimes.com/blogs/voices/how-home-healthcare-can-help-patients-with-serious-kidney-ailments-keep-their-condition-under-check/

[12] C. Clinic, "Kidney failure," accessed on April 29, 2024. [Online]. Available: https://my.clevelandclinic.org/health/diseases/17689-kidney-failure

[13] A. S. Levey, D. Cattran, A. Friedman, W. G. Miller, J. Sedor, K. Tuttle, and ..., "Proteinuria as a surrogate outcome in ckd: report of a scientific workshop sponsored by the national kidney foundation and the us food and drug administration," *American Journal of Kidney Diseases*, vol. 54, no. 2, pp. 205–226, 2009.

[14] J. R. Chapman, "What are the key challenges we face in kidney transplantation today?" *Transplantation Research*, vol. 2, no. 1, pp. 1–7, 2013.

[15] S. Gerogianni, F. Babatsikou, G. Gerogianni, E. Grapsa, G. Vasilopoulos, S. Zyga, and C. Koutis, "'concerns of patients on dialysis: A research study'," *Health Science Journal*, vol. 8, no. 4, p. 423, 2014.

[16] J. R. Chapman, "What are the key challenges we face in kidney transplantation today?" *Transplantation Research*, vol. 2, no. 1, pp. 1–7, 2013.

[17] F. Ma, T. Sun, L. Liu, and H. Jing, "Detection and diagnosis of chronic kidney disease using deep learning-based heterogeneous modified artificial neural network," *Future Generation Computer Systems*, vol. 111, pp. 17–26, 2020.

[18] M. H. Hesamian, W. Jia, X. He, and P. Kennedy, "Deep learning techniques for medical image segmentation: achievements and challenges," *Journal of digital imaging*, vol. 32, pp. 582–596, 2019.

[19] N. Bhaskar and M. Suchetha, "A computationally efficient correlational neural network for automated prediction of chronic kidney disease," *IRBM*, vol. 42, no. 4, pp. 268–276, 2021.

[20] V. Singh, V. K. Asari, and R. Rajasekaran, "A deep neural network for early detection and prediction of chronic kidney disease," *Diagnostics*, vol. 12, no. 1, p. 116, 2022.

[21] M. Majid, Y. Gulzar, S. Ayoub, F. Khan, F. A. Reegu, M. S. Mir, W. Jaziri, and A. B. Soomro, "Enhanced transfer learning strategies for effective kidney tumor classification with ct imaging," *International Journal of Advanced Computer Science and Applications*, vol. 14, no. 8, 2023.

[22] S. Sudharson and P. Kokil, "An ensemble of deep neural networks for kidney ultrasound image classification," *Computer Methods and Programs in Biomedicine*, vol. 197, p. 105709, 2020.

[23] D.-H. Kim and S.-Y. Ye, "Classification of chronic kidney disease in sonography using the glcm and artificial neural network," *Diagnostics*, vol. 11, no. 5, p. 864, 2021.

[24] E.-H. A. Rady and A. S. Anwar, "Prediction of kidney disease stages using data mining algorithms," *Informatics in Medicine Unlocked*, vol. 15, p. 100178, 2019.

[25] M. Bhandari, P. Yogarajah, M. S. Kavitha, and J. Condell, "Exploring the capabilities of a lightweight cnn model in accurately identifying renal abnormalities: Cysts, stones, and tumors, using lime and shap," *Applied Sciences*, vol. 13, no. 5, p. 3125, 2023.

[26] A. Bhattacharjee, S. Rabea, A. Bhattacharjee, E. B. Elkaeed, R. Murugan, H. M. R. M. Selim, R. K. Sahu, G. A. Shazly, and M. M. Salem Bekhit, "A multi-class deep learning model for early lung cancer and chronic kidney disease detection using computed tomography images," *Frontiers in Oncology*, vol. 13, p. 1193746, 2023.

[27] S. Kanwal, M. A. Khan, A. Fatima, M. M. Al-Sakhnini, O. Sattar, and H. Alrababah, "Ia2skabs: Intelligent automated and accurate system for classification of kidney abnormalities," in *2022 International Conference on Cyber Resilience (ICCR)*. IEEE, 2022, pp. 1–10.

[28] S. Wasi, S. B. Alam, R. Rahman, M. A. Amin, and S. Kobashi, "Kidney tumor recognition from abdominal ct images using transfer learning," in *2023 IEEE 53rd International Symposium on Multiple-Valued Logic (ISMVL)*. IEEE, 2023, pp. 54–58.

[29] K. Yildirim, P. G. Bozdag, M. Talo, O. Yildirim, M. Karabatak, and U. R. Acharya, "Deep learning model for automated kidney stone detection using coronal ct images," *Computers in biology and medicine*, vol. 135, p. 104569, 2021.

[30] F. Ma, T. Sun, L. Liu, and H. Jing, "Detection and diagnosis of chronic kidney disease using deep learning-based heterogeneous modified artificial neural network," *Future Generation Computer Systems*, vol. 111, pp. 17–26, 2020.

[31] N. Heller, F. Isensee, K. H. Maier-Hein, X. Hou, C. Xie, F. Li, Y. Nan, G. Mu, Z. Lin, M. Han *et al.*, "The state of the art in kidney and kidney tumor segmentation in contrast-enhanced ct imaging: Results of the kits19 challenge," *Medical image analysis*, vol. 67, p. 101821, 2021.

[32] S. Srivastava, R. K. Yadav, V. Narayan, and P. K. Mall, "An ensemble learning approach for chronic kidney disease classification," *Journal of Pharmaceutical Negative Results*, pp. 2401–2409, 2022.

[33] N. Bhaskar and S. Manikandan, "A deep-learning-based system for automated sensing of chronic kidney disease," *IEEE Sensors Letters*, vol. 3, no. 10, pp. 1–4, 2019.

[34] F. P. Schena, V. W. Anelli, J. Trotta, T. Di Noia, C. Manno, G. Tripepi, G. D'Arrigo, N. C. Chesnaye, M. L. Russo, M. Stangou *et al.*, "Development and testing of an artificial intelligence tool for predicting end-stage kidney disease in patients with immunoglobulin a nephropathy," *Kidney international*, vol. 99, no. 5, pp. 1179–1188, 2021.

[35] G. Chen, C. Ding, Y. Li, X. Hu, X. Li, L. Ren, X. Ding, P. Tian, and W. Xue, "Prediction of chronic kidney disease using adaptive hybridized deep convolutional neural network on the internet of medical things platform," *IEEE Access*, vol. 8, pp. 100 497–100 508, 2020.

[36] M. N. Islam, M. Hasan, M. K. Hossain, M. G. R. Alam, M. Z. Uddin, and A. Soylu, "Vision transformer and explainable transfer learning models for auto detection of kidney cyst, stone and tumor from ct-radiography," *Scientific Reports*, vol. 12, no. 1, p. 11440, 2022.

[37] C.-C. Kuo, C.-M. Chang, K.-T. Liu, W.-K. Lin, H.-Y. Chiang, C.-W. Chung, M.-R. Ho, P.-R. Sun, R.-L. Yang, and K.-T. Chen, "Automation of the kidney function prediction and classification through ultrasound-based kidney imaging using deep learning," *NPJ digital medicine*, vol. 2, no. 1, p. 29, 2019.

[38] M. N. Islam, M. Hasan, M. K. Hossain, M. G. R. Alam, M. Z. Uddin, and A. Soylu, "Vision transformer and explainable transfer learning models for auto detection of kidney cyst, stone and tumor from ct-radiography," *Scientific Reports*, vol. 12, no. 1, p. 11440, 2022.

[39] M. Tan and Q. V. Le, "Efficientnetv2: Smaller models and faster training," 2021.

[40] C. Szegedy, V. Vanhoucke, S. Ioffe, J. Shlens, and Z. Wojna, "Rethinking the inception architecture for computer vision," 2015.

[41] M. Sandler, A. Howard, M. Zhu, A. Zhmoginov, and L.-C. Chen, "Mobilenetv2: Inverted residuals and linear bottlenecks," 2019.

[42] A. Dosovitskiy, L. Beyer, A. Kolesnikov, D. Weissenborn, X. Zhai, T. Unterthiner, M. Dehghani, M. Minderer, G. Heigold, S. Gelly, J. Uszkoreit, and N. Houlsby, "An image is worth 16x16 words: Transformers for image recognition at scale," 2021.

[43] F. Zabihollahy, N. Schieda, S. Krishna, and E. Ukwatta, "Automated classification of solid renal masses on contrast-enhanced computed tomography images using convolutional neural network with decision fusion," *European Radiology*, vol. 30, pp. 5183–5190, 2020.

[44] D. Akgün, A. T. KABAKUŞ, Z. K. ŞENTÜRK, A. ŞENTÜRK, and E. Küçükkülahli, "A transfer learning-based deep learning approach for automated covid-19diagnosis with audio data," *Turkish Journal of Electrical Engineering and Computer Sciences*, vol. 29, no. 8, pp. 2807–2823, 2021.

[45] M. Kalkan, G. E. Bostancı, M. S. Güzel, B. Kalkan, Ş. Özsarı, Ö. Soysal, and G. Köse, "Cloudy/clear weather classification using deep learning techniques with cloud images," *Computers and Electrical Engineering*, vol. 102, p. 108271, 2022.

[46] S.-W. Lee, "Novel classification method of plastic wastes with optimal hyperparameter tuning of inception_resnetv2," pp. 274–279, 2021.